*Review*

# Applications of Recurrent Neural Network for Biometric Authentication & Anomaly Detection

Joseph M. Ackerson, Rushit Dave * and Naeem Seliya

Department of Computer Science, University of Wisconsin Eau-Claire, Eau Claire, WI 54701, USA; ackersjm6535@uwec.edu (J.M.A.); seliyana@uwec.edu (N.S.)
* Correspondence: daver@uwec.edu

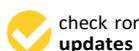

**Abstract:** Recurrent Neural Networks are powerful machine learning frameworks that allow for data to be saved and referenced in a temporal sequence. This opens many new possibilities in fields such as handwriting analysis and speech recognition. This paper seeks to explore current research being conducted on RNNs in four very important areas, being biometric authentication, expression recognition, anomaly detection, and applications to aircraft. This paper reviews the methodologies, purpose, results, and the benefits and drawbacks of each proposed method below. These various methodologies all focus on how they can leverage distinct RNN architectures such as the popular Long Short-Term Memory (LSTM) RNN or a Deep-Residual RNN. This paper also examines which frameworks work best in certain situations, and the advantages and disadvantages of each proposed model.

## 1. Introduction

People have always been fascinated by the idea of creating an artificial human brain and these efforts became known as artificial neural networks (ANN). ANNs are hardly a novel concept, but the numerous ways in which they have been applied are revolutionizing the world. There are numerous variations of specialized ANNs; take convolutional neural networks (CNN), for example, which are adapted to work specifically with image or video data. This paper focuses specifically on the applications of Recurrent Neural Networks (RNN). RNNs are unique because they are comprised of many neural networks chained together, which allows them to process a series of data where a network learns from its previous experiences. RNNs have a wide array of applications, ranging from written language to speech recognition.

Security for our devices and data is of increasing concern in recent years. RNNs have the potential to improve upon current methods, but also allow advancements in new authentication techniques. Biometric authentication usually relates to a phone sensor that can read a fingerprint or iris. These are things often found in a modern smart phone. However, biometric authentication is so much more than that. What if it were feasible to use biometric authentication to protect cloud data in transit from a mobile device [1]? This opens new avenues for the application of biometric authentication. A few examples of biometric authentication are mouse movement authentication, keystroke authentication [2], handwritten password authentication [3], and even palm print authentication [4,5]. Moving away from sensor-based biometric authentication makes it available to numerous different uses that previously required a specific sensor. Not only will this allow for more accessible biometric authentication, but it will keep the system and devices more secure as these types of biometrics are much harder to impersonate. RNNs can also open the environments in which authentication is performed.

Another key implementation of Recurrent Neural Networks is in the field of facial recognition. Facial recognition ranges from identifying one's identity to deciphering their emotions. Expression recognition often relies on a CNN for extraction of important features from image data before that image data can be used by the RNN [6]. Once these features are deciphered the LSTM RNN can make a prediction about the emotion perceived. Emotional recognition is important for many reasons, especially with the rapid development of robotics. The ability for software to be able to distinguish different human emotions will be of increasing importance in the future. Emotional and expression recognition will increase acceptance and help dissolve the barrier of interactions between man and machine.

One popular implementation of RNNs is applied to the domain of anomaly detection. Anomaly detection can range from detecting spam emails, to malicious network traffic and maritime vessel traffic. Anomaly detection can also be utilized in aviation [7]. The application of RNNs to the field of aviation is relatively new. These specialized neural networks can help detect anomalous flight conditions, predict excessive engine vibrations, determine the remaining life of a turbine engine, and aid in landing [8]. Anomaly detection is important for maintaining safety and security in many aspects of everyday life. It looks at which patterns are normal and denotes an event outside of the margin of normal operation as anomalous. One such application of anomaly detection can be applied to Internet of Things (IoT) devices. IoT devices can include smart speakers, thermostats, and even fridges. The goal of the paper [9] is to detect patterns in IoT devices which can then be applied to track unusual patterns in a network of IoT devices. An example of Anomaly Detection in IoT devices can be seen in [10] where researchers develop an Intrusion Detection System (IDS) for IoT devices. An IDS using a RNN would rely on detecting anomalous patterns in the data to alert a user if there was anyone trying to hack into their IoT devices.

These are the four main topics that this paper will be reviewing. The goal of this paper is to analyze novel approaches in each of the four applications of RNNs. The remainder of this paper is organized as follows: background discussion of current research, review of biometric authentication, review of facial recognition, review of anomaly detection and aircraft, discussion and analysis of each topic covered in the literature review, discussion and analysis, limitations, conclusion, and future work.

## 2. Background

Authenticating users to ensure they are the ones who are truly accessing their data has been a difficult task for as long as computers have been around. Password authentication has been and still is one of the most popular ways to verify a user is who they say they are. However, passwords have too many flaws and are often not unique to one person. The best form of authentication is one in which the user can utilize something unique to them. This is where we get biometric authentication, as these biometric systems provide an alternative approach to authenticate whereby physiological or behavioral characteristics are sensed and used to authenticate a user. Physical biometrics use features about a person like an iris, fingerprint, or face. However, these types of biometrics require an expensive fingerprint or iris sensor. An alternative would be behavior biometrics such as electrocardiogram (ECG) signals, mouse and keyboard patterns, and handwriting patterns. Utilizing these behaviors based biometric systems, biometric authentication can be more widely available and will not rely on expensive sensors in devices. RNNs can also help to improve upon current password and sensor based biometric authentication methods. This allows access to new environments where biometric authentication previously was unavailable. Paper [11] aims to use biometric authentication for patient identification in hospitals. This type of system would be especially useful in a situation with an unresponsive patient, as doctors would be able to find the patient's medical information based on their biometric data alone. An additional new environment that is important for novel authentication techniques is in IoT devices [12]. This is another area in which the application of RNN-based biometric authentication can be implemented. A group of researchers are exploring breathing acoustics-based authentication for IoT devices [13]. This adds biometric authentication that is natural to a user, making authentication simple and not something the user needs to think about day to day.

Authentication is a common target for malicious intent, and biometric authentication is not as secure as most users believe it is. These methods can be breached very quickly, as attackers adapt as quickly as new security innovations are released. Since biometric data is so unique to the individual, losing it can be far more detrimental than getting one password stolen. Based on the current technology, there is no replacing stolen fingerprint data. So, there need to be new methods of biometric authentication that do not require something physical, but rather some mental behavior or pattern. This is where the RNNs have the potential to dramatically improve how biometric authentication is performed and improve upon current sensor-based authentication methods. This can be seen in [14], where researchers authenticated based on eye movement patterns. RNNs perform best with time-series data, which allows multiple neural networks to work together to verify the identity of a user. This could mean scanning your fingerprint multiple times or tracking your mouse and identifying patterns in the movement. Nevertheless, RNNs do have advantages and disadvantages in authentication [15] and this is important to continue exploring to improve biometric authentication techniques.

Facial recognition implementations can also reap rewards of advancements in RNN research. Facial recognition can be a tough, but important topic to discuss in today's societal atmosphere. It is a technology that is used for surveillance; however, it has other applications which need continued research. This paper focuses on a sub-category of facial recognition, which concentrates on analyzing human facial expressions. "Human emotion recognition is a challenging machine learning task with a wide range of applications in human-computer interaction, e-learning, health care, advertising, and gaming" [16]. Expression recognition is an essential technique to improve interactions between humans and machines. This is especially important in the field of robotics, as it will allow robots to understand and differentiate between different emotions and adjust its interactions accordingly. Expression recognition works by using a CNN to analyze the video input, which gets passed to an RNN for analysis at each time step to determine the emotion occurring in each frame. Then, a final prediction is made about the facial expression seen in the video clip.

Currently, robots only understand how to behave based on their programming and are not very adaptable to the person interacting with them. Facial expression recognition can change these impersonal interactions entirely. "Emotion can reflect information of hobbies, personality, interests and even health, recognition of human emotions can help machines and robots in improving the reliability of human-machine interaction" [17]. In addition to distinguishing emotions at a basic level, robots will be capable of reproducing facial expressions. Creating more "human" robots will be a vital step in allowing for the looming integration of robots into everyday life to happen smoothly.

Having a group of people look over log files or sift through data trying to find anomalies is a very insufficient solution. This is another field where recent advancement in RNN research can make a big improvement. There is such a wide range of possibilities to apply anomaly detection. Common applications can already be observed in our everyday lives, as seen through detection of spam emails, combing through network traffic logs to find attackers, and even real-time flight data analysis. RNNs are a great "alternative approach to cyber-security detection frameworks" [18]. Anomaly detection had the ability to prevent incidents from happening using RNNs to detect issues before they became a major problem. Today, there are many instances in which people do not know if what they are reading is true. This can lead to many consequences, some of which are already unfolding. Anomaly detection can read through data before it becomes trending and determine if it is real or fake. Have you ever wondered if the product review you read on a webpage is real or just fake reviews manufactured by bots? Again, this is an avenue in which anomaly detection can help spot fake information. These are just a few common issues facing society today, and with continued development in RNNs, there is a chance to combat them.

Appling anomaly detection techniques to aviation is a rapidly growing practice. Anomaly detection in aviation can range from diagnosing excessive engine vibration to determining the remaining lifespan of jet engines. Continued development will allow for improved safety of flights as well as a deeper understanding of aviation. Aircraft maintenance is one such sector where RNNs are making many improvements. RNNs can predict when certain parts need maintenance or need to be replaced altogether. This will help streamline the maintenance process and ensure less downtime for aircraft.

## 3. Literature Review

### 3.1. Novel Smartphone Authentication Techniques

Sensors such as iris scanners or fingerprint readers are amongst the most popular forms of smartphone biometric authentication. RNNs can improve upon sensor-based approaches by not only improving existing methods, but also by opening opportunities to develop new sensor-based biometric authentication methods. One novel approach to biometric authentication is through inertial gait recognition [19]. "Fingerprint and face recognition is based on a physical characteristic, but biometrics can also recognize how a user performs a specific activity" [19]. Inertial gait recognition is based upon walking patterns detected in an individual, making authentication not only easier but something that can be done without even having to think about it. The purpose of this paper is to explore a novel approach to biometric authentication through inertial gait recognition. The model that was developed involves data taken from a gyroscope and accelerometer. These values are processed into gait signals and then fed into an RNN. This proposed model for inertial gait recognition can be seen in Figure 1. The OUDB database was selected to train and evaluate this model. The OUDB consists of two datasets, one measured on a flat surface and another on a sloped surface, with a total of 744 male and female users of varying ages [19]. Many different RNN models were tested with varying vector size, number of filters, and fully connected layers. The best results came from a Filter size of 64, 2 fully connected layers, and a vector size of 128. This proposed method had a training/testing Equal Error Rate (EER) of 11.48%/7.55% respectively. Other novel approaches to smartphone authentication are through ECG signals [20,21] and holding position combined with touch type authentication [13]. The combined accuracy of smartphone hand position and touch-typing [13] detection leads to an accuracy of 93.9% with the proposed model. Models from [14,15] both utilized the specialized LSTM cell. Using this LSTM cell, the ECG signal-based authentication reached accuracies of 100% [20] for using the MITDB dataset and 99.73% [21].

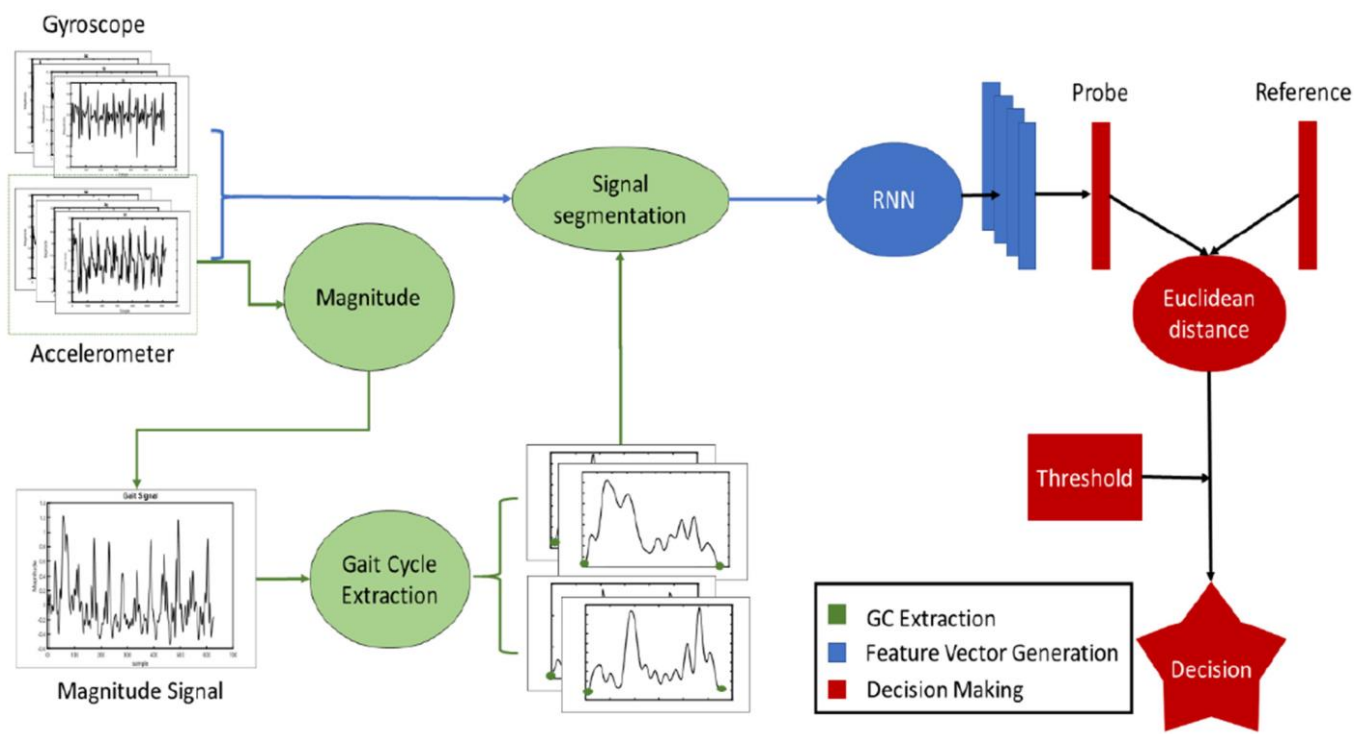

**Figure 1.** Proposed model for inertial gate authentication [19].

*3.2. Mouse and Keyboard Based Authentication Methods*

An increasingly popular form of biometric authentication is through the recognition of mouse movements or keyboard-based behavioral patterns. Rapid User Mouse Behavior Authentication (RUMBA) [22] is a novel attempt to detect patterns in mouse movements using RNNs and the architecture of this model is represented in Figure 2. The researchers took this approach because monitoring physical characteristics requires access to extra hardware like specialized sensors. The paper also describes that data like mouse movement information is easy to collect and contains little privacy-sensitive information. The proposed method involves a fusion of a CNN-RNN, since complex identification tasks benefit from utilizing the fusion of two types of neural networks. To test this CNN-RNN neural network the researchers used a database provided by the Xi'an Jiaotong University of China. The dataset consists of 15 users, each completing 300 trials. The goal was to click on static targets around the screen 8 times per trial. The best results came from the fusion CNN-RNN model [23], which was able to authenticate users with an accuracy of 99.39%. Similar techniques to a mouse-based approach are keystroke-based authentication systems, which are the focus of [24–26]. Paper [24] uses a simple LSTM based structure to detect keystroke dynamics and evaluates this model using a dataset from Carnegie Mellon University. This dataset comes from 51 users, measuring the times it took them to enter a password, and time in between individual letter or symbol keypresses. The results of [24] reached 100% accuracy after 1500 epochs. The results of [18] using the UMDAA-02 dataset and LSTM RNN architecture. The model [25] reduced to an ERR of 19% when fusing all modalities. Another group of researchers [26] used a CNN-RNN based approach to authenticate users based on keystroke data. They trained and tested this model using the SUNY Buffalo dataset which contains 157 participant's fixed and free text data. The proposed model [26] was able to obtain a final EER of 3.04%.

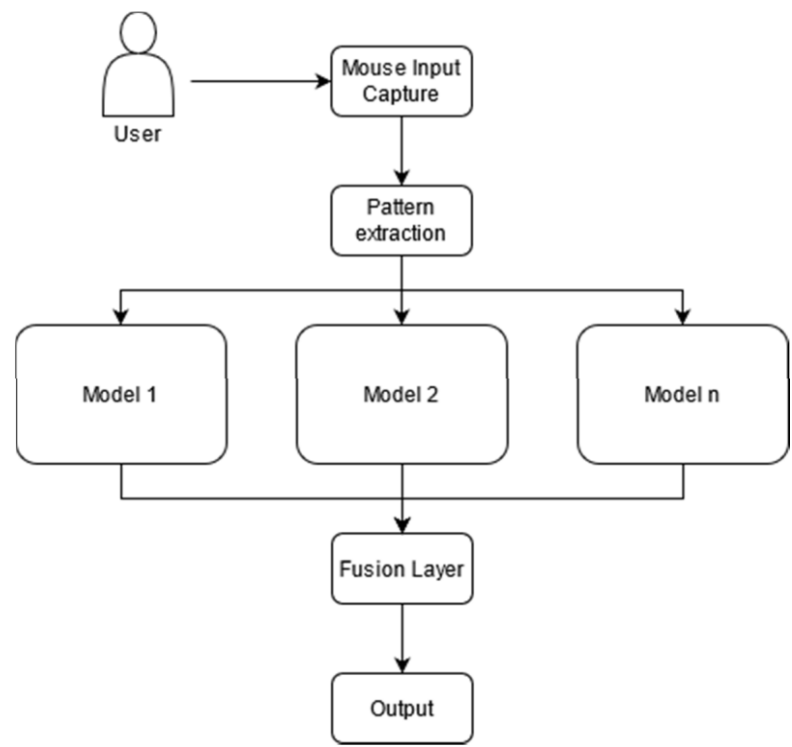

**Figure 2.** Proposed model for mouse behavior authentication.

*3.3. Handwritten Authentication Methods*

A person's handwriting is a unique and distinguishable trait no matter how neat or messy it is. Handwriting-based authentication methods aim to determine a user's identity based on how they write. One such method directly implies an LSTM RNN to analyze a user's signature which is also represented in Figure 3. Their proposed method uses Siamese architecture [27]. This model is then trained and tested with the BiosecurID database. This database is comprised of 16 signatures and 12 professional forgeries per user with a total of 400 total users. The researchers also gathered X and Y pen coordinates, pressure, and timestamp using a pen tablet. When this data is fed into the LSTM network the final EER was 6.44% for 1:1 and 5.58% for 4:1 (ratio of number of original signatures to skilled forgeries). These results prove that this methodology [27] would be an even lower EER with random or unskilled forgeries. Another attempt to authenticate users from their fingerprint data uses handwritten passwords instead of a signature. This would be like drawing each digit of a 4-letter pin code [28]. The methodology is similar to that of the previous example, except these researchers use a bidirectional LSTM network after Siamese architecture. To train and evaluate their model, these researchers created their own dataset by the name of e-BioDigit. Their dataset is composed of online handwritten digits from 0–9. To collect this data, each user would use their fingers to write out the digits 0–9 a total of four times over two sessions. Using this dataset, the proposed method [28] was able to accurately authenticate with an EER of 3.8%.

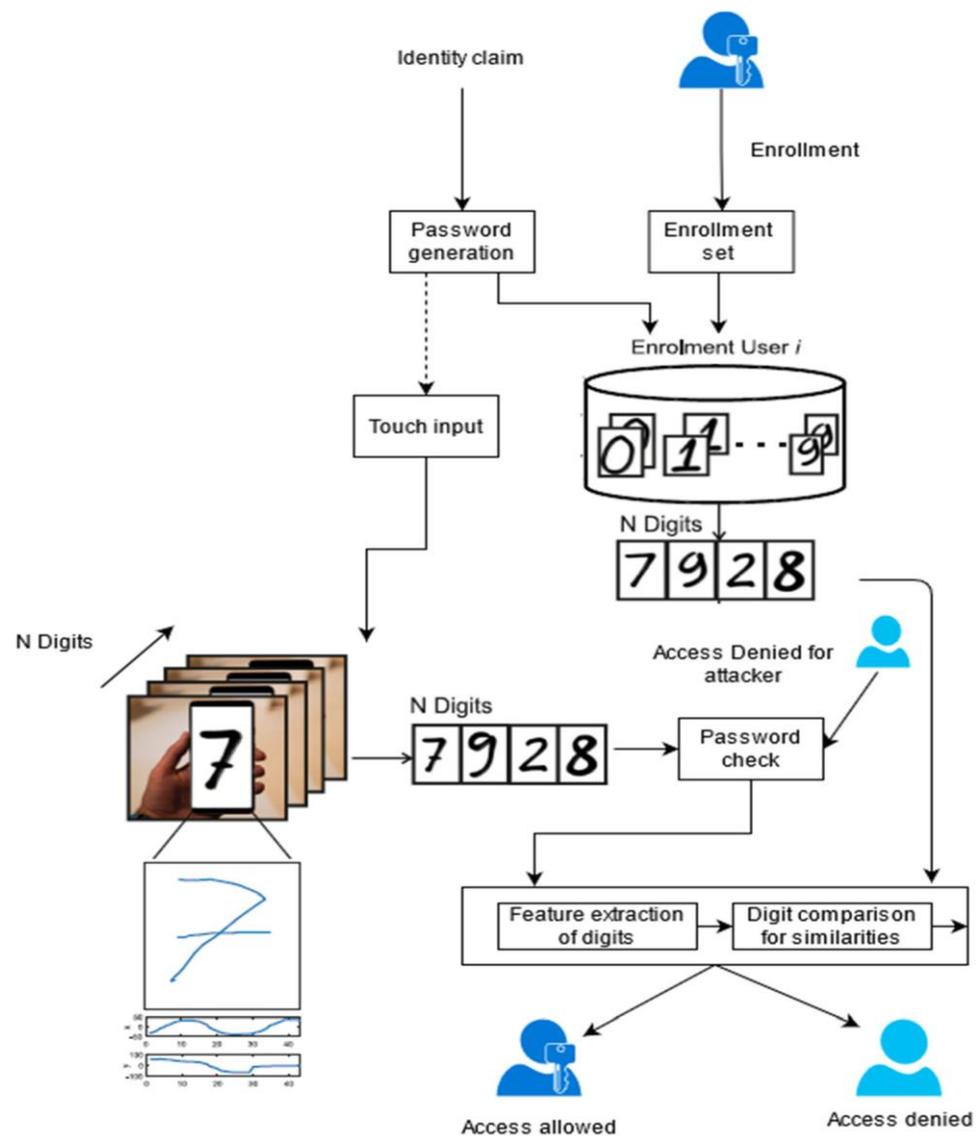

**Figure 3.** Proposed model for handwritten authentication.

*3.4. Model for Facial Expression Recognition Using LSTM RNN*

Facial Expression recognition has been a popular task, one which is also benefiting from the use of an LSTM RNN. This paper [29] feeds a dual CNN structure into an LSTM RNN gate, which can be seen in Figure 4, to process the extracted features from the video frame. These researchers choose to use four different datasets to train and test their model. These datasets are the extended Cohn-Kanade database, which contains 593 image sequences from 123 different subjects, the MMI dataset, which consists of 2885 videos of facial expression from 88 subjects, the Static Facial Expressions in the Wild dataset, which is made up of 663 expression samples, and finally their own dataset, compiled from 80 subjects who each performed the 6 basic emotions. The six basic emotions present in each of these datasets are fear, disgust, anger, happiness, sadness, surprise, and neutral. With their proposed method [29], they were able to attain 99% on CK + dataset, 81.60% on MMI, 56.68% on SFEW (which is highly accurate for that dataset), and 95.21% on their own dataset. Other similar methodologies [30,31] were also able to benefit from the LSTM gate implemented in their models and were evaluated against the MMI dataset. The model from [30] was able to achieve an impressive accuracy of 92.07%, and the proposed method from [31] attained an accuracy of 82.97%.

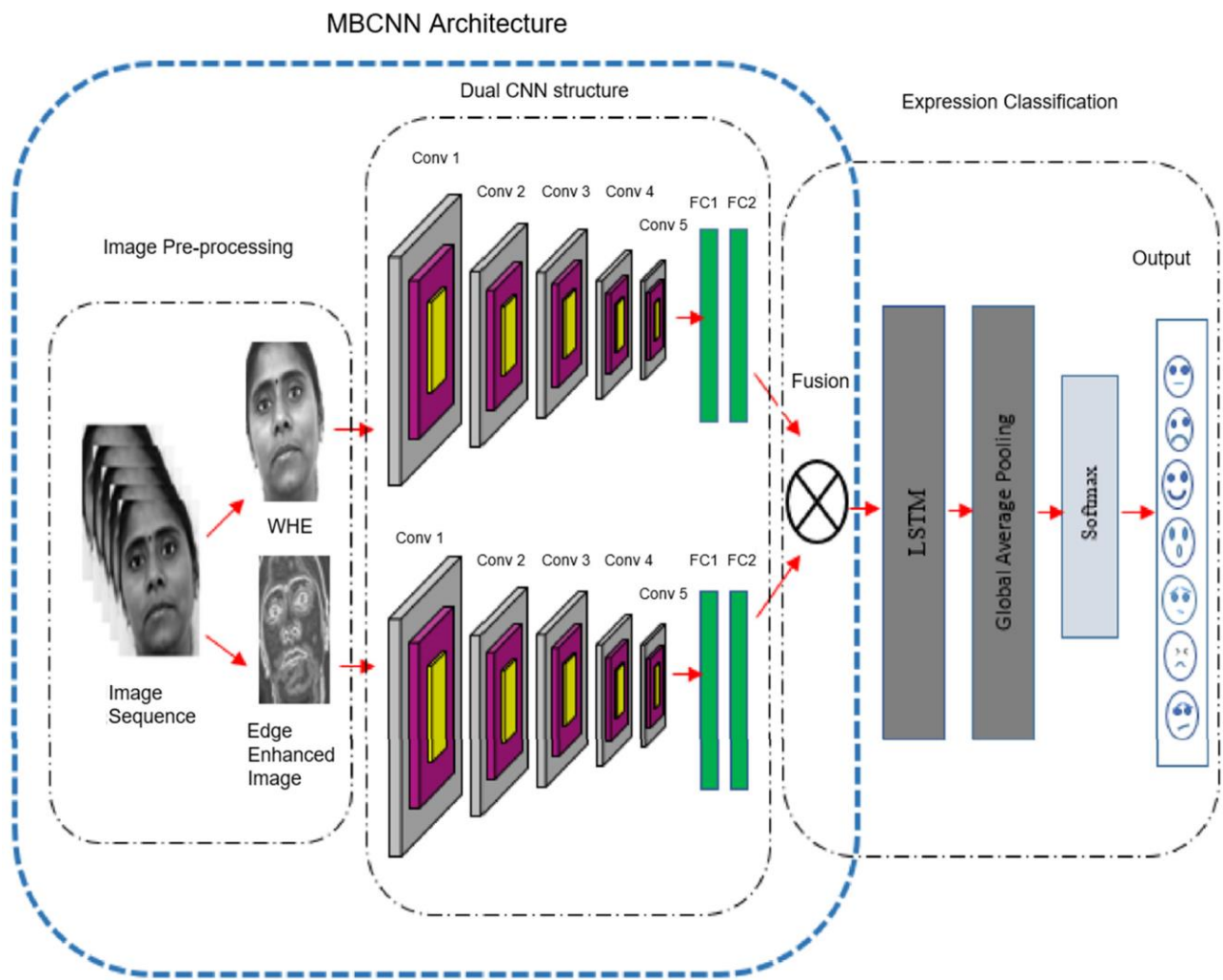

**Figure 4.** Proposed model for CNN+LSTM authentication [29].

*3.5. Multimodal Expression Recognition Implementing an RNN Approach*

　　The multimodal approach to expression recognition implements multiple modalities into the RNN framework to improve recognition accuracy. These types of modalities include, but are not limited to, facial expressions, speech, head movements, and body movements. All these traits help to determine someone's feelings and emotions. Having input from multiple modalities can be confusing since the computer must make sense of these different inputs. So, feature extraction is of the upmost importance to ensure an accurate prediction. The dataset that was applied to this model [32] was the AVEC2015 dataset, which is a section taken from the RECOLA dataset. This dataset contains modalities like audio, video, electrocardiogram, and electrodermal activity for each subject, with the emotions of arousal and valence being portrayed. The best results from this proposed model [32] were divided, with the best arousal results coming from the early fusion of all the modalities into the LSTM network that is displayed in Figure 5, and the best valence results coming from the late fusion methodology. Both strategies, however, combine all the different modalities into the LSTM RNN structure, allowing them the best Root Mean Squared Error or RMSE. Another group's proposed method [33] was able to achieve similar results to the previous model using the same dataset as shown in Figure 6.

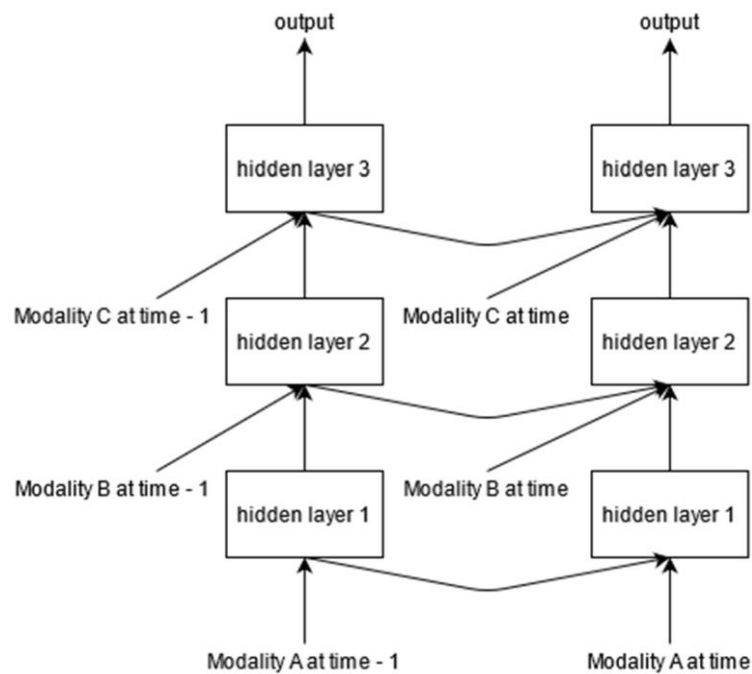

**Figure 5.** Proposed model for multimodal expression recognition.

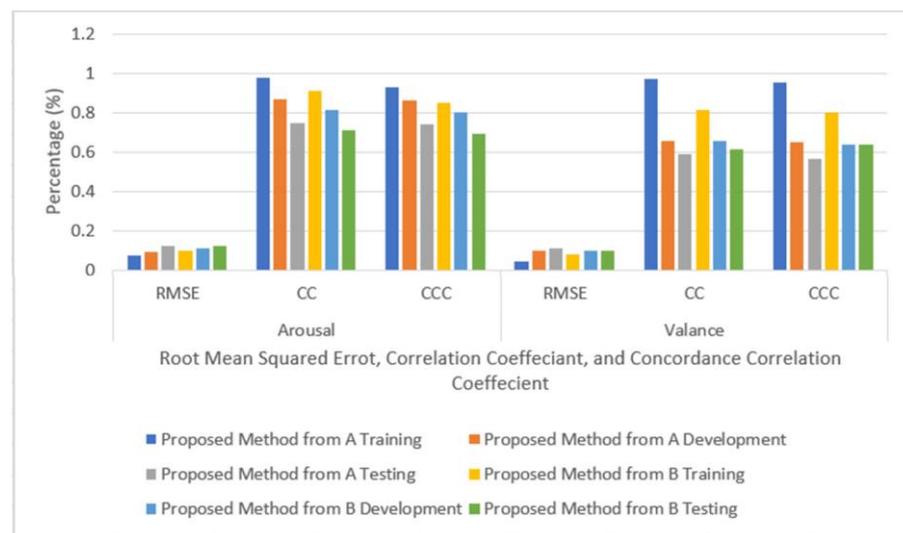

**Figure 6.** Root Mean Squared Error, Correlation Coefficient, and Concordance Correlation Coefficient.

*3.6. Motion History Image Expression Recognition*

A Motion History Image (MHI) is an image that has a record of all movements in a single image. The method from this paper [34] utilizes Locally Enhanced MHI to extract features to pass and fuses this with a Cross Temporal Segment LSTM RNN shown in Figure 7. This type of fusion layer was able to reach an accuracy of 93.9% on the CK + dataset. Their model was also evaluated against the MMI and AFEW datasets, where the model [34] was able to achieve an accuracy of 78.4% and 51.2% respectively. Extracting these temporal features was also the goal of [35]. This proposed model extracts the temporal geometry and spatial features, then fuses them to be passed into the LSTM RNN. Using this methodology, this model [35] was able to evaluate facial expressions at an accuracy of 81.71% against the MMI dataset beating. Both models [34,35] surpass methods that rely solely on a CNN to detect expression. This is where an LSTM becomes helpful in extracting temporal features.

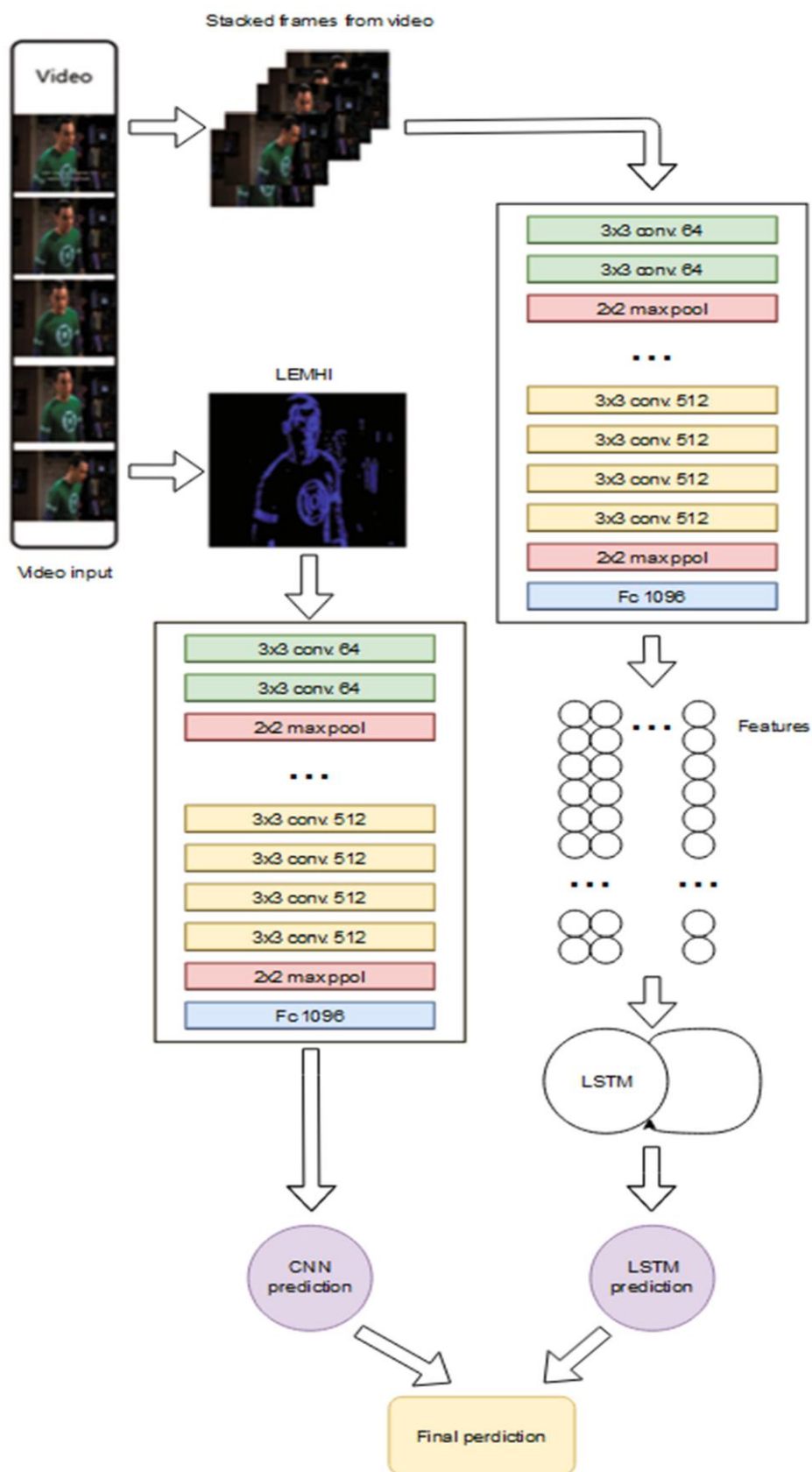

**Figure 7.** Proposed model for motion history image.

### 3.7. Anomaly Detection of Maritime Vessels

The goal of the research done in paper [36] is to improve transportation and shipping through anomaly detection to increase awareness of all vessels and reduce potential accidents. The researchers use an LSTM RNN architecture to track anomalous vessel movements by feeding it trajectory data shown in Figure 8. The RNN will use this trajectory data to determine if the vessel has shifted from the next tracking point and decide if this is anomalous. The data comes from the algorithm, or Density-Based Spatial Clustering of Applications with Noise (DBSCAN), which is used to determine these tracking points. The dataset used to train and test this model was gathered from an Automatic Identification System from one of the largest ports in the word located in China's Zhoushan Islands. The RNN was able to detect anomalous course, speed, and route. In this case, the course is the current trajectory, and the route is the total path to the destination. The network caught each instance the vessel was behaving irregularly. The anomaly can also be applied to occupancy detection, anomalous exchange rate prices, network anomaly detection, and anomalous stock price detection. The researchers tested multiple different models of [37] and evaluated their model against all four of these potential situations. The results can be seen in Figure 9.

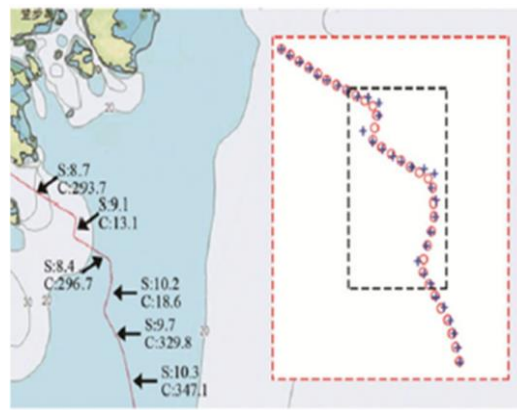

**Figure 8.** Anomalies in vessels' course [36].

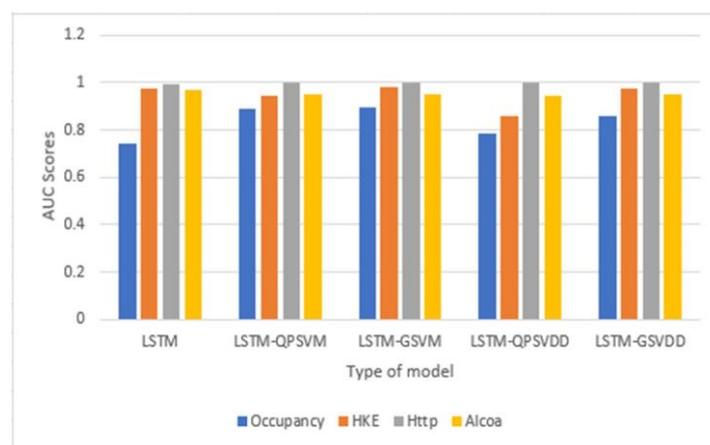

**Figure 9.** Area under Curve for multiple LSTM based models.

### 3.8. Anomaly Detection in Water Quality

Regulating and monitoring water quality is important for the health and safety of all who rely on that water supply. With a RNN and a dataset collected from real world data [38], it is possible to monitor the quality of water flowing through a water treatment facility. The structure of the RNN is shown in Figure 10. This dataset was collected from a public water company in Germany, by the name of Thüringer Fernwasserversorgung.

This data consists of temperature, chlorine dioxide levels, acidity (pH) levels, etc. Using this dataset to train and evaluate an LSRM RNN, the model [38] was able to achieve and F1 score of 0.9023. LSTM RNNs can also be used for anomaly detection in network traffic. The methodology of [39] uses TCP dump data collected over 5 weeks to train and test the model. This model [39] was able to reach an accuracy of 94% while only triggering 2 false alarms, 98% while triggering 16 false alarms, and 100% while triggering 63 false alarms.

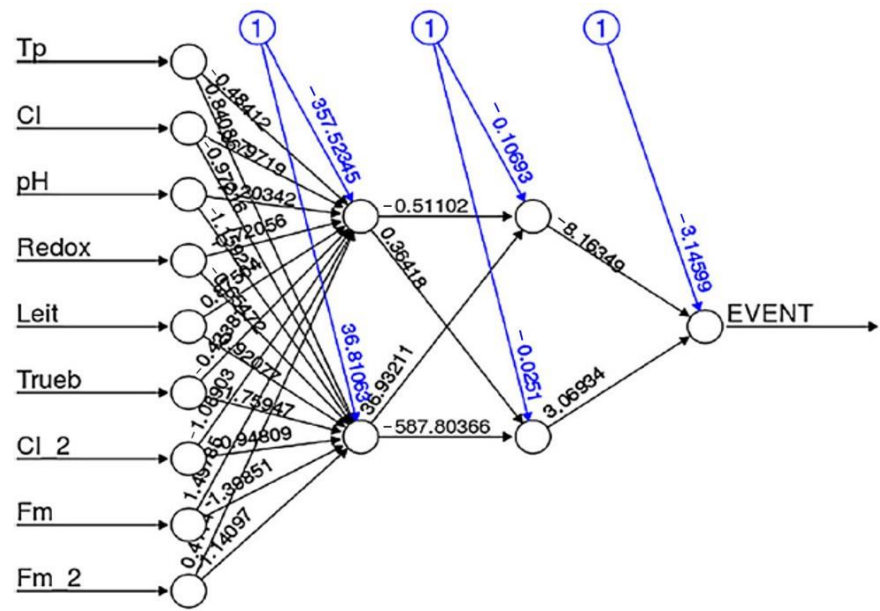

**Figure 10.** Proposed model for water quality anomalies [31].

*3.9. Stacked RNN Strategy for Anomaly Detection in Pedestrian Areas*

Anomaly detection can also apply to tracking and identifying abnormal occurrences surrounding events such as running, loitering, or driving. The framework of a stacked RNN (sRNN) involves stacking multiple RNNs, represented in Figure 11, on top of each other, as done in [40]. This sRNN was evaluated against four different databases, being CUHK Avenue, USCD Pedestrian 1 and 2, Subway, and their custom dataset. Each dataset is comprised of multiple videos displaying normal and abnormal events. An example of an abnormal event would be when a car drives in an area where there are usually pedestrians. The sRNN can go frame by frame through these videos and track the anomaly as it progresses through the scene. Using the sRNN, the architecture [40] was able to achieve accuracies of 81.71% on CUHK Avenue, 92.21% on Pedestrian 2, and 68.00% on their custom dataset. RNN based strategies can also be useful for detecting anomalies in network traffic. Another model [34] that is using an RNN attempts to detect cyber-attacks against Supervisory Control and Data Acquisition (SCADA) equipment in an industrial plant. The model was evaluated against generated data using the Tennessee Eastman Process (TEP). The results of the implementation [41] used the Numenta Anomaly Benchmark to get a score of 0.373 and achieved a score of 0.823 with the DoS attack type.

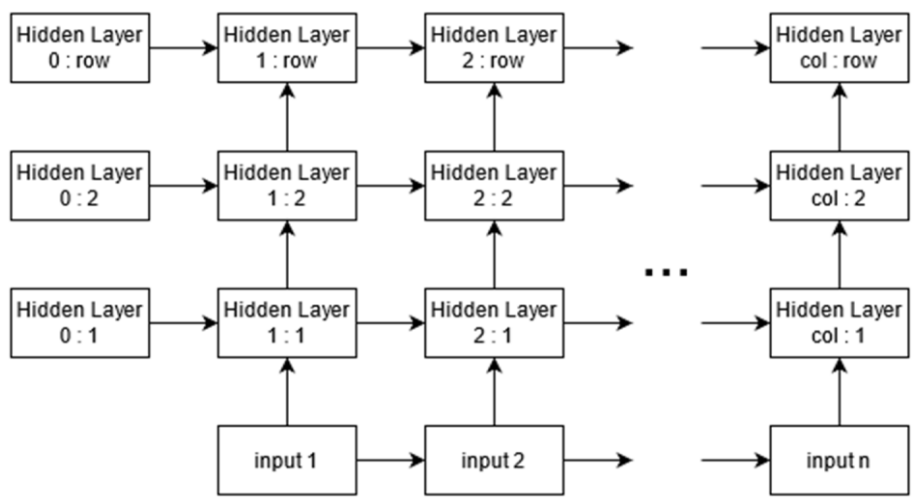

**Figure 11.** Proposed model for stacked RNN.

*3.10. Physics Based Aircraft Flight Trajectory Prediction*

Flight trajectory prediction is an important tool for planning and executing a safe flight from one destination to another. The methodology behind [42] is to use a physics-based approach to reduce the cost of simulating aircraft trajectories, which can be very computationally expensive. This type of cost increases even further when multiple aircraft trajectories need to be simulated in real time. This method aims to cut down the cost of these simulations using a Deep Residual RNN (DR-RNN) which is compared to a data-based LSTM RNN simulation technique. The architecture of an LSTM-RNN for predicting flight trajectory can be seen in Figure 12. The data used to evaluate both approaches was based on a Boeing 747-100 cruising at 40,000 feet. The DR-RNN was able to accurately match its predictions within an indistinguishable error rate. In case 2, or longitudinal responses, the prediction error was $3.20 \times 10^{-7}$, and in case 3, or lateral responses, the prediction error was $1.17 \times 10^{-5}$ [42]. The LSTM approach was close to where the predictions of the DR-RNN were, but it had a more difficult time making accurate predictions, whereas the DR-RNN's predictions are always in line with the true values. A different LSTM based approach to flight trajectory prediction [43] uses data collected from Automatic Dependent Surveillance-Broadcast (ADS-B stations). These ADS-B stations transmit aircraft positional information with high accuracy. This data was collected over a period of 5 months. The model [43] was able to lower MRSE to 0.2295, 0.1337, and 123.512 for latitude, longitude, and height, respectively.

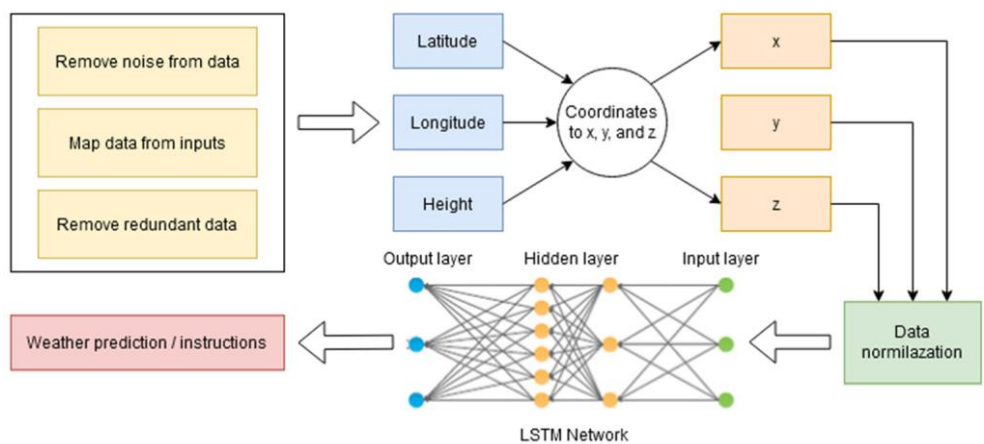

**Figure 12.** Proposed model for aircraft dynamics simulation.

### 3.11. Real Time Anomaly Detection Onboard Unmanned Aerial Vehicles

Detecting anomalies in flight patterns of an Unmanned Aerial Vehicle or UAVs is important for maintaining a higher rate of reliability and safety. The methodology of [44] explores the possibilities of applying an LSTM RNN, as shown in Figure 13, using real sensor data from a UAV flight to validate the model. The data from the flight is collected and the network is trained with normal flight data. For evaluation of the model, point anomalies are introduced into the flight data. The types of introduced anomalies are in the UAVs forward velocity and pneumatic lifting. This proposed method [44] was able to reach an accuracy of 99.7% for forward velocity anomalies and 100% for pneumatic lifting anomalies. A similar methodology can be applied to detecting anomalies in manned aircraft, specifically commercial airline flights. The data used to construct the model [45] was gathered from a C919 airliner belonging to Commercial Aircraft Corporation of China (COMAC). During a test flight the researchers were able to gather terabytes of sensor data. This model [45] was able to achieve an accuracy of 99.4% based on the confusion matrix. Researchers were able to improve upon a similar model to [45] in [46] by using Field Programmable Gate Array acceleration. An FPGA accelerated LSTM RNN was able to perform at a speed of 28.76 times faster than the CPU against the same COMAC's dataset. Another group of researchers [47] also tried to detect anomalous flight parameters using data generated by X-Plane simulations. Using these simulations, the researchers were able to simulate data from 500 flights, 485 of which were normal and 15 of which were anomalous. Types of anomalies that were being detected were very high airspeed approach, landing runway configuration change, influence of wind, high airspeed for short duration, etc. The proposed LSTM RNN model [47] was able to get an F1 score of 0.899.

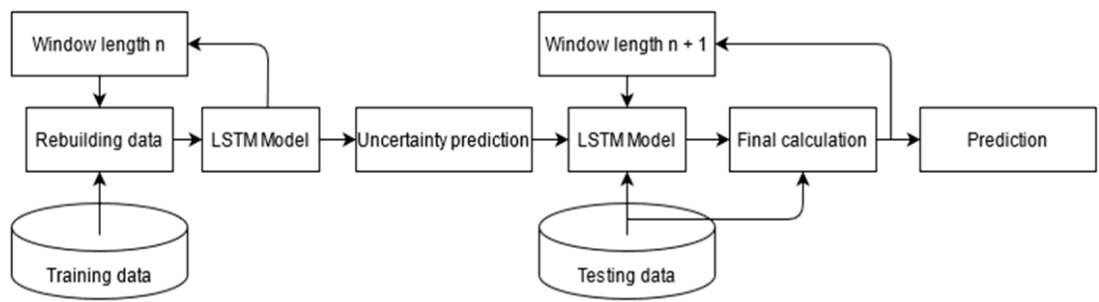

**Figure 13.** Proposed model for UAV anomaly detection.

### 3.12. Prediction of Remaining Life of Jet Turbine Engines

Being able to predict how much longer a jet engine will last can, not only increase the safety of pilots and passengers, but also ensure these engines are being used to the fullest extent and are properly maintained along the way. The methodology of [48] is to use a LSTM-HMM fusion architecture, which can be seen in Figure 14, to predict remaining engine life. To evaluate, train, and test this model, researchers used simulated data from Commercial Modular Aero-Propulsion System Simulation (C-MAPSS). The C-MAPSS system simulated an engine at 90,000 pounds of thrust at different altitudes from sea level to 40,000 ft, Mach 0 to 0.90, and sea-level temperatures from −60 to 103 degrees Fahrenheit. This data was fed into the LSTM-HMM network, and the model was able to achieve an F1 score of 0.781. This is an improvement from the LSTM only model [48], which got an F1 score of 0.715. LSTM RNNs can also be used to detect excess engine vibration. If a turbine engine has excess vibrations, it can advise engineers that an engine needs maintenance or replacement. Recognition of these access engine vibrations was the goal of [49,50]. The method of [49] was to use 15 different parameters recorded by Flight Data Recorder (FDR). These parameters are altitude, angle of attack, bleed pressure, turbine inlet temperature, mach Number, etc. These parameters were taken from a subset of 76 parameters captured from the FDR when a flight suffered from excess vibrations. The purpose of this model was to predict engine vibrations. Three different LSTM architectures were tested, and the best results achieved errors rates (MAE) of 0.033048, 0.055124, and 0.1011991 at 5, 10 and 20 s, respectively. The method of [50] is to use ant colony optimization on the LSTM from [49]. This optimization improved the MAE of the 10 s prediction from 0.055124 to 0.0427.

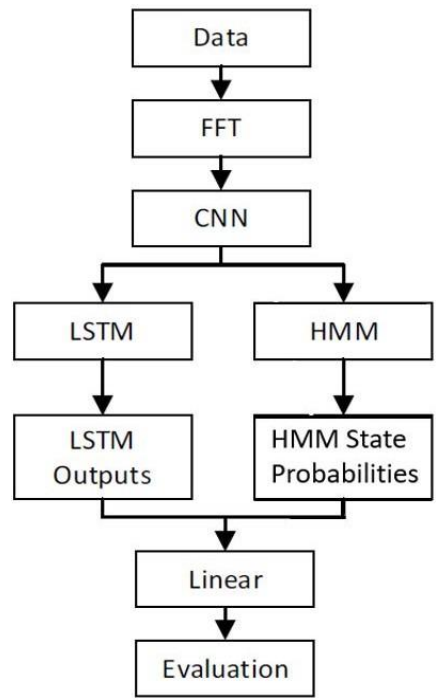

**Figure 14.** Proposed model for remaining life engine prediction [48].

## 4. Discussion and Analysis

In Table 1 above each of the three main papers from all four topics are summarized by methodology which includes the structure and data collection strategies, the results of each of the papers along with the dataset used and inference time if available, and finally the pros and cons of each paper. Each method of biometric authentication discussed above has a unique application and one might want to choose a method to better fit their needs, for example the mouse movement authentication technique can be a very simple, portable, and secure method. However, a drawback is that it may take longer for users to configure their information when compared to a fingerprint reader or take less thought like inertial gait authentication. For any authentication technique, there is always a balance speed and security.

Choosing the best method for facial expression recognition might be slightly more straight forward since you would like a method that is both fast and accurate. All the papers reviewed above had great scores, but image processing still takes the most time depending on the pixel density of each frame in the video and given that a 3 s video at 60 fps is 180 frames that need to be propagated through the network.

Anomaly detection is another area where the application or where you are looking for anomalies matters. RNNs have proven that they work well in analyzing and detecting anomalies in time series data and should be recommended based on the results above. Now there are different types of RNNs like an LSTM-RNN or a stacked RNN framework and this is there the application will determine what type of architecture is the most appropriate. A growing category of anomaly detection is in aviation. Aviation is a newer and growing section of anomaly detection that focuses on all parts of the aircraft from engine vibration to its trajectory. An RNN based approach has also been proven to be the most useful strategy in aviation as well and any new models would greatly benefit from an LSTM-RNN approach if there is any trouble on deciding what model to use.

Table 1. Comparison of existing RNN applications.

| Title | Methodology | Results | Pros and Cons |
|---|---|---|---|
| Novel Smartphone Authentication Techniques [19] | Using an RNN to authenticate users through inertial gait recognition or identify users based on their physical movement patterns. Gait recognition also requires gyroscope and accelerometer sensor data to track movement. | The best performing results obtained an equal error rate of 11.48% using 20% for training, and 7.55% using 70% for training. These results were obtained from the Osaka University Database (OUDB). | Users can authenticate based on walking patterns. Makes authentication easier, allowing it a wider range of applications. However, sensors are required to collect inertial gait data. |
| Mouse and Keyboard Based Authentication Methods [23] | Authenticate uses a CNN+RNN fusion to detect behavioral patterns in mouse movement. All this requires is a mouse and a program that can capture the mouse input data. | The proposed model was able to accurately authenticate users 99.39% of the time. The dataset for this paper was provoided by Xi'an Jiaotong University of China. | Sensors are not required for biometric authentication; all you need is a mouse. However, authentication could take longer as you may need to perform a longer process to authenticate. |
| Handwritten Authentication Methods [27] | Employing an LSTM RNN to analyze users' handwriting and confirm or deny them access to a system. To collect user data, there needs to be some sort of device like a tablet for users to write write their signature. | The LSTM RNN was able to achieve a final EER of 6.44% for 1:1 and 5.58% 4:1. 1:1 and 4:1 are the ratios of real signatures to skilled forgeries. These researchers generated their own development and evaluation datasets. Each training iteration lasted approximatley 30 min with 200 training iterations and 100 testing iterations. | This has more potential than entering a password as it adds an extra layer of security to passwords. However, having to handwrite passwords requires some sort of device or touch screen. |
| Model for Facial Expression Recognition Using LSTM RNN [29] | Utilizing an LSTM RNN for facial expression recognition against multiple datasets including one developed by the researchers. A camera is needed to collect the neccesary video data used to build an expression recognition dataset. | The LSTM RNN was able to reach an accuracy of 99% on CK+ dataset, 81.60% on MMI dataset, 56.68% on SFEW dataset, and 95.21% on their own dataset. | The LSTM RNN has shown great promise in expression recognition. However, feature extraction methodologies can hinder recognition accuracies. |
| Multimodal Expression Recognition Implementing an RNN Approach [32] | Multimodal expression recognition uses multiple modalities like speech, body movement, head movement, etc. All these elements are combined to recognize emotions. Since multiple modalities need to be collected, one might need a camera for video, micophone for audio, and possibly body tracking sensor data. | Results can be seen in Figure 9. The dataset used for this challenge (AVEC2015) was a subset of the larger RECOLA dataset. | Multiple modalities can improve upon expression recognition. However, extracting and processing all these different features can create lots of noise and cause inaccuracies. |
| Motion History Image Expression Recognition [34] | Using a Cross Temporal LSTM RNN with Motion History Images for facial expression recognition. Motion History images are essentially multiple images stacked on top of each other to form one image that shows a sequence of events so no additional sensors are required. | The proposed method was able to achieve an accuracy of 93.9%, 78.4%, and 51.2% on CK+, MMI, and AFEW datasets, respectively. | Motion History Images allow for all motion in a video to be captured on a still image, leading to easier feature extraction. However, these images can often be cluttered, creating a lot of noise. |

Table 1. *Cont.*

| Title | Methodology | Results | Pros and Cons |
| --- | --- | --- | --- |
| Anomaly Detection of Maritime Vessels [36] | Leveraging an RNN to detect anomalies in course, speed, and trajectory of vessels with density-based clustering. Vessel data was gathered from the Automatic Identification Ststem (AIS). | The results can be seen in Figure 10. The dataset for this model was built from the DBSCAN algorithm which was applied to AIS data to generate trajectory points used to train the network. | There is a lot of traffic to sort through within busy ports which, if handled correctly, can significantly improve accuracy. However, there exists room for false positives when dealing with large volumes of data. |
| Anomaly Detection in Water Quality [38] | Using an RNN to monitor the quality of, and detect anomalous traits in, water flowing through a control facility in Germany. Additional sensors needed to measure water quality would contain instruments to measure temperature, acidity, and chlorine dioxide levels plus any other water quality traits. | The proposed model was able to achieve an F1 score of 0.9023. The reseachers built their own dataset from a real sensor data taken from Thüringer Fernwasserversorgung public water company. | A high F1 score means not many false alarms were triggered. However, monitoring multiple different qualities in water can make triggering a false positive or false negative more common. |
| Stacked RNN Strategy for Anomaly Detection in Pedestrian Areas [40] | Applying a stacked RNN framework to detect anomalous events and activities in pedestrian areas. For this paper researchers used data collected from cameras that were in public areas. | Using the sRNN, the model was able to achieve accuracies of 81.71% on CUHK Avenue, 92.21% on Pedestrian 2, and 68.00% on their custom dataset. The sRNN model takes about one hour to train on the Avenue dataset and takes 0.02 s to make a prediction from any frame. | Stacked RNN frameworks provide many different cells both vertically and horizontally. However, with this type of anomaly detection it is hard to define what events are anomalies. |
| Physics Based Aircraft Flight Trajectory Prediction [42] | Utilizing a Deep Residual RNN to predict the flight trajectory of aircraft and reduce computational cost of aircraft simulations. This DR-RNN was compared to a more typical LSTM RNN. A tool would be needed to create flight simulations and be able to gather that simulation data. | In case 2, or longitudinal responses, the prediction error was $3.20 \times 10^{-7}$. In case 3, or lateral responses, the prediction error was $1.17 \times 10^{-5}$. The dataset used to train the DR-RNN was gathered from simulated data of a Boeing 747-100 with introduced anomalies. | Deep Residual RNNs allow for the integration of aircraft dynamics into the simulations used to calculate aircraft trajectories. Both models outperformed previous numerical based simulation methods. |
| Real Time Anomaly Detection Onboard Unmanned Aerial Vehicles [44] | Leveraging and LSTM RNN to detect real time flight data anomalies in UAV drone data. Sensors onboard the drone were able record and log data during the drones flights. | This proposed model was able to get an accuracy of 99.7% for forward velocity anomalies, and 100% for pneumatic lifting anomalies. The dataset used in this model came from the actual flight data logged by drone flights. | Detecting anomalies in real-time can be difficult when using an LSTM RNN architecture. The data coming off the UAV will need to be forwarded through the network quickly to constantly ensure the drone is operating properly. |
| Prediction of Remaining Life of Jet Turbine Engines [48] | The researchers devised a fusion network built from an LSTM-HMM to predict remaining life of a jet turbine engine. Data was gathered from 21 sensors outside and inside of the jet turbine engine to measure vibrations. | The LSTM-HMM network scored an F1 accuracy of 0.781. The dataset used to train and evaluate this model came from the C-MAPSS dataset. | There is often a lot of noise within data coming from engine sensor data, aking sure excess vibration anomalies are being correctly identified can be difficult. |

Recurrent Neural Networks have many benefits over other styles of machine learning methods. RNN's have the unique ability for each cell to have its own memory of all the previous cells before it. This allows for RNN's to process sequential data in time steps which other machine learning models cannot do. Think about teaching a computer to read a single word. How will it know what the word is if it is always forgetting the previous letters that it has seen so far? For some application of machine learning like identifying an image or finding patterns in static data an RNN would not be necessary. However, when you want to do speech recognition, auto generation of captions, or even having a computer generate music, it needs to hold on to that sequential data to help predict the next state. Common applications of RNN's one can find in everyday life is any voice assistant available on your phone i.e., Google or Alexa. Call centers can take advantage of RNN's to handle basic support tasks taking the burden off human operators. RNN's can also be found sorting through your emails to sort out spam and phishing emails from friendly emails. All the applications that have been discussed above have also all seen an improvement when applying a RNN based learning model for their chosen application. Not only have they seen a benefit over previous machine learning models, but RNN's also open more possibilities for new ways in which machine learning can accomplish a certain task.

## 5. Limitations

Recurrent Neural Networks show that they are up to the task of solving many issues with a high rate of success. However, they are not perfect and require future research to improve upon existing research. RNNs are still just proving to be able to bring new possibilities to biometric authentication, expression recognition, anomaly detection, and aviation. These applications are still in their infancy and require continued research to improve accuracy and precision. The novel research and models shown in this paper have displayed great potential but come with their own issues. Authentication models sometimes struggle to authenticate uses under certain contexts, and sensors have potential to fail. Facial recognition models struggle under certain lighting conditions, which can cause inaccurate recognition. Anomaly detection methods can trigger false alarms and sometimes miss an anomalous event. Models used for aircraft recognition struggle when there is too much noise in the data, which can lead to inaccurate predictions.

## 6. Conclusions and Future Work

The goal of this paper was to provide insights into current research being done in four similar yet very distinct fields. These areas are biometric authentication, expression recognition, anomaly detection, and aviation. Each paper reviewed has been pushing the limits and striving to bring new and exciting innovations to their respective areas of research. This paper specifically looked at how Recurrent Neural Networks were changing the game and allowing for new innovations. With continued research into these areas, there can be even more improvement in each of these areas: making sure that user data and critical systems are secured with top-level biometric authentication, paving a road for improvement in interactions between man and machine, detecting malicious actors and making sure people stay safe through novel anomaly detection techniques, and making air travel even safer while getting the most use out of aircraft parts. Future work done in these fields should push to improve upon the current models that have been reviewed here and should work to develop novel methodologies of their own.

**Funding:** This research received no external funding.

**Data Availability Statement:** Data sharing not applicable.

**Conflicts of Interest:** The authors declare no conflict of interest.